\pdfoutput=1

\documentclass[11pt]{article}

\usepackage[preprint]{acl}

\usepackage{times}
\usepackage{latexsym}
\usepackage{amsmath}
\usepackage{colortbl}
\definecolor{ForestGreen}{RGB}{34,139,34}
\usepackage{multirow}
\usepackage{array}
\usepackage{graphicx}
\usepackage{subcaption}
\usepackage[T1]{fontenc}

\usepackage[utf8]{inputenc}

\usepackage{microtype}

\usepackage{booktabs}
\usepackage{inconsolata}

\usepackage{graphicx}

\title{
MCTS-Judge: Test-Time Scaling in LLM-as-a-Judge for \\Code Correctness Evaluation}

\author{Yutong Wang \\ Mechanical Engineering \\ National University of Singapore \\[0.3cm] {\bf Kaixin Li} \\ School of Computing \\ National University of Singapore
         \And
        Pengliang Ji \\ Robotics Institute \\ Carnegie Mellon University \\[0.3cm]
		{\bf Ming Hu} \\ Electrical and Computer Systems \\ Monash University \\[0.3cm]
		{\bf Guillaume Sartoretti} \\ Mechanical Engineering \\ National University of Singapore \\
		\small{e0576114@u.nus.edu, guillaume.sartoretti@nus.edu.sg}
        \And
        Chaoqun Yang \\ Independent Contributor \\ $\,$ \\[0.3cm]
		{\bf Jiaoyang Li} \\ Robotics Institute \\ Carnegie Mellon University 
}

\begin{document}

\maketitle
$\,$ \\[1cm]

\begin{abstract}
The \emph{LLM-as-a-Judge} paradigm shows promise for evaluating generative content but lacks reliability in reasoning-intensive scenarios, such as programming.
Inspired by recent advances in reasoning models and shifts in scaling laws, we pioneer bringing test-time computation into LLM-as-a-Judge, proposing \textbf{MCTS-Judge}, a resource-efficient, System-2 thinking framework for code correctness evaluation.
MCTS-Judge leverages Monte Carlo Tree Search (MCTS) to decompose problems into simpler, multi-perspective evaluations.
Through a node-selection strategy that combines self-assessment based on historical actions in the current trajectory and the Upper Confidence Bound for Trees based on prior rollouts, MCTS-Judge balances global optimization and refinement of the current trajectory.
We further designed a high-precision, unit-test-level reward mechanism to encourage the Large Language Model (LLM) to perform line-by-line analysis.
Extensive experiments on three benchmarks and five LLMs demonstrate the effectiveness of MCTS-Judge, which improves the base model's accuracy from $41.0\%$ to $80.0\%$, surpassing the o1-series models with $3 \times$ fewer tokens. 
Further evaluations validate the superiority of its reasoning trajectory in logic, analytics, thoroughness, and overall quality, while revealing the test-time scaling law of the LLM-as-a-Judge paradigm.
\end{abstract}

\section{Introduction}
\textit{LLM-as-a-Judge}, wherein Large Language Models (LLMs) serve as the golden rule for evaluation criteria~\cite{gu2024survey}, has been proposed for applications such as generative content assessment~\cite{li2024salad}, and data captioning~\cite{chen2024mllm}, serving as a cost-effective solution compared to human expert evaluators.
Among those, LLM-as-a-Judge has revolutionized code evaluation by automating judgment~\cite{yang2024evaluating}, repair~\cite{liu2024marscode}, and explanation~\cite{weyssow2024codeultrafeedback}, replacing inaccurate similarity-based execution-free methods~\cite{ren2020codebleu, tran2019does}, and expensive execution-based methods reliant on manually-crafted test cases~\cite{zheng2023codegeex, zhuo2024bigcodebench}.

\begin{figure}
    \centering
	\vspace{2cm}
    \includegraphics[width=0.7\linewidth]{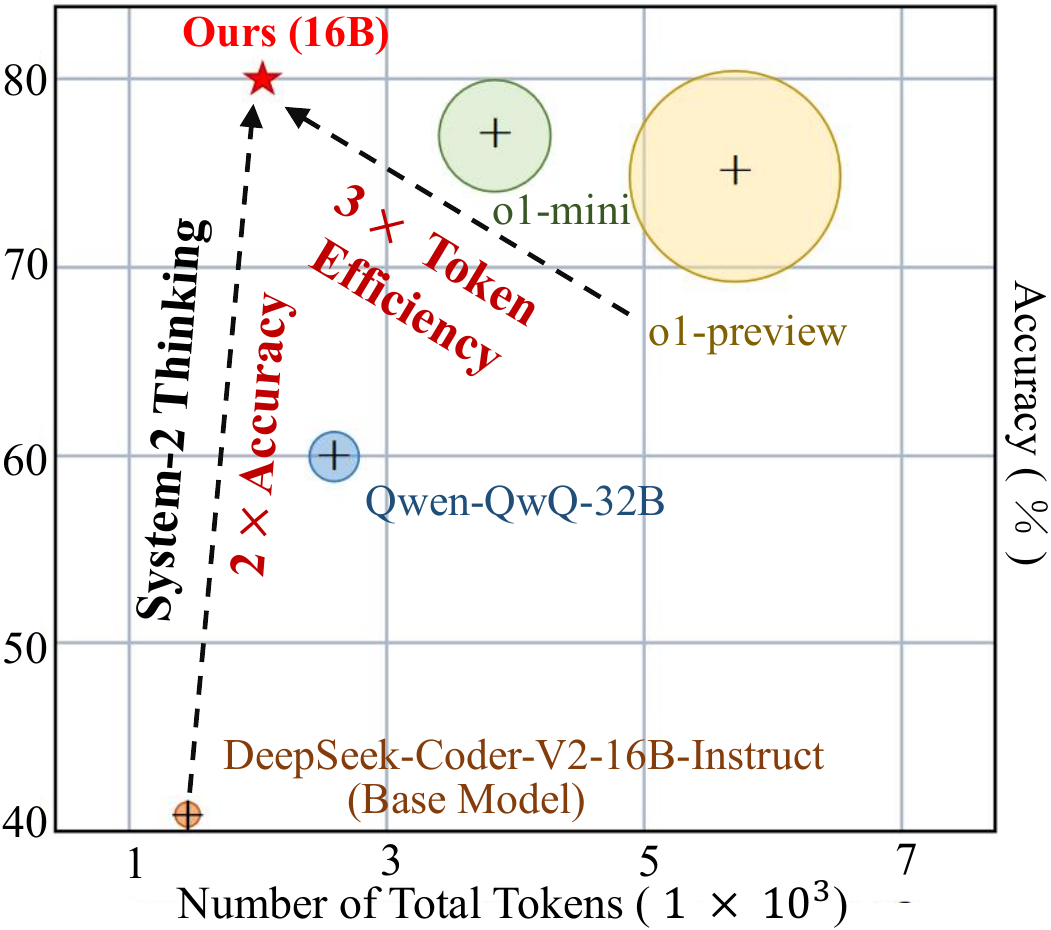}
    \vspace{-0.2cm}
    \caption{
    With test-time scaling, our MCTS-Judge method doubles the accuracy of \textit{DeepSeek-Coder-V2-16B-Instruct} on the APPS benchmark, surpassing o1-series models and \textit{Qwen-QwQ-32B}, while using $3 \times$ fewer tokens and a smaller model.
    The circle sizes indicates the relative sizes of the models.}
    \label{fig:teasor}
    \vspace{-0.6cm}
\end{figure}

Despite its growing adoption, recent studies highlight critical challenges in the LLM-as-a-Judge paradigm, including bias~\cite{gu2024survey}, misalignment~\cite{ye2024justice}, and fairness concerns~\cite{li2024llms}, questioning its reliability for accurate, human-like judgments. 
To address these issues, researchers have focused on pretraining~\cite{hui2024qwen2}, fine-tuning~\cite{wang2024enhancing}, and in-context learning~\cite{wei2022chain} to improve reasoning capabilities, which are highly demanded in programming scenarios.
Unfortunately, as LLMs near the upper bounds imposed by scaling laws, further advancements involve increasing costs in training with diminishing returns~\cite{snell2024scaling}. 

To address these limitations, inspired by the shift of scaling laws from training to test time~\cite{xu2025towards} and recent breakthroughs in \textit{Reasoning LLMs}, such as OpenAI's o-series~\cite{jaech2024openai}, we introduce the first framework that integrates test-time computation into the LLM-as-a-Judge paradigm.
We target code correctness evaluation and propose MCTS-Judge, a resource-efficient LLM-as-a-Judge framework with System-2 thinking, offering human-like reasoning for more reliable evaluations. 
It achieves State-Of-The-Art (SOTA) performance compared to prior LLM-as-a-Judge methods, which rely on rapid and superficial System-1 thinking~\cite{tong2024codejudge, zhuo2023ice}.
MCTS-Judge leverages a tailored Monte Carlo Tree Search (MCTS) to decompose problems into simpler, multi-perspective evaluation tasks.
In the selection phase of MCTS, we introduce a global-local node selection strategy that combines self-assessment based on historical actions in the current trajectory, and the Upper Confidence Bound for Trees (UCT) algorithm, guided by prior rollouts, to balance the optimization of high-value regions in the global search space with local reasoning trajectories.
We further designed a high-precision simulated execution reward mechanism. 
This mechanism combines cost-effective automatic test case synthesis with LLM-as-an-interpreter execution, prompting line-by-line analysis for unit-test-level reliability.

Extensive experiments on five LLMs across three challenging code benchmarks, including BigCodeBench~\cite{zhuo2024bigcodebench}, HumanEval-X~\cite{zheng2023codegeex}, and APPS~\cite{hendrycks2021measuring}, with varying code complexity and languages, highlight the reliability of MCTS-Judge powered by test-time computation. 
As shown in Fig.~\ref{fig:teasor}, our approach elevates the accuracy of \textit{DeepSeek-Coder-Lite-16B}~\cite{zhu2024deepseek} from 41.0\% to 80.0\%, surpassing o1-series models~\cite{jaech2024openai} and open-source Qwen-QwQ-32B~\cite{qwen2024technical}, while using only 3$\times$ fewer tokens and a smaller model. 
Furthermore, we achieve SOTA performance on all experiments compared to previous System-1 thinking-based LLM-as-a-judge frameworks, with up to 32\% improvement on APPS, and demonstrate strong robustness in generalizable scenarios without code references.
Case studies on HumanEval-X further showcase MCTS-Judge's superior reasoning across four fine-grained dimensions, such as logic and analytics, achieving a higher win rate over o1-series models. 
Finally, we validated that scaling test-time computation, including tree depth and rollouts, further enhances MCTS-Judge's accuracy, shedding light into the test-time scaling law for LLM-as-a-judge paradigms.

\section{Related Work} 
\subsection{Code Correctness Evaluation} 
Code correctness evaluation can be broadly broken down into two paradigms.
Execution-free methods, such as BLEU~\cite{papineni2002bleu}, ROUGE-L~\cite{lin2004rouge}, METEOR~\cite{denkowski2014meteor}, ChrF~\cite{popovic2015chrf}, RUBY~\cite{tran2019does}, and CodeBLEU~\cite{ren2020codebleu}, assess code based on textual or code-specific feature similarity to reference code.
In this paper, we refer to them as similarity-based evaluation methods. 
However, reference code is often unavailable in practice, and these methods struggle to distinguish semantically equivalent but syntactically different code, leading to low accuracy, as shown in Appendix~\ref{appendix:comparision-exec}.
In contrast, execution-based methods, commonly used in code generation benchmarks~\cite{zheng2023codegeex, zhuo2024bigcodebench}, assess code correctness by executing it against test cases. 
However, this approach demands comprehensive handcrafted test cases and isolated environments, making it costly and operationally complex~\cite{chen2022codet}.
To address these limitations, recent efforts have explored LLM-as-a-Judge paradigms with in-context learning. ICE-Score~\cite{zhuo2023ice} integrates evaluation criteria into prompts, while CODEJUDGE~\cite{tong2024codejudge} employs a two-stage prompting approach. However, these methods rely on System-1 thinking~\cite{kahneman2011thinking}, leading to rapid, superficial decisions that are constrained by the inherent uncertainties of LLMs, resulting in limited reliability.

\subsection{Test-time Computation Boost Reasoning}
Recent studies highlight a shift in scaling laws from train-time to test-time~\cite{ji2025test, xu2025towards}, as pretrained models approach data scale limits~\cite{snell2024scaling}, while reasoning models leverage test-time computation, demonstrating remarkable performance improvements, exemplified by OpenAI's o-series models~\cite{jaech2024openai}.
To advance human-like System-2 thinking, key innovations include chain-of-thought data curation~\cite{wang2022self, wang2024strategic}, reinforcement learning~\cite{deepseek2025r1, qwen2024technical}, and reward models~\cite{guan2025rstar, yu2024self}.
As a core support, search paradigms like beam search and MCTS dynamically select diverse reasoning trajectories, significantly enhancing accuracy in large search spaces. Examples include ReST-MCTS~\cite{zhang2024rest}, rStar~\cite{qi2024mutual}, MCTSr~\cite{zhang2024accessing}, and~\cite{xie2024monte}, which integrate MCTS with reinforced self-training, self-play mutual reasoning, and preference optimization, driving advancements in reasoning tasks such as math and code problem-solving.
Building on this remarkable improvement in reliability, we pioneeringly integrate test-time computation into the LLM-as-a-Judge paradigm, proposing a novel framework, MCTS-Judge, which leverages System-2 thinking to generate reliable, human-like reasoning trajectories for comprehensive, multi-perspective code correctness evaluation.

\section{MCTS-Judge} 
In this section, we first introduce the overview of MCTS-Judge for code evaluation (Sec.\ref{Sec3.1}), then detail its MCTS architecture (Sec.\ref{Sec3.2}) and reward mechanism (Sec.~\ref{Sec3.3}).

\subsection{Overview}
\label{Sec3.1}
The code correctness evaluation task determines whether a code snippet $c$ correctly implements the functionality described in a problem statement $p$, expressed as $x = (c, p)$. 
In MCTS-Judge, we decompose this task into subtasks, each prompting the LLM to verify a specific requirement.
The action space of our MCTS consists of these subtasks and a null action representing no evaluation.
At each node in the search tree, the subaction space includes one non-repeating subtask and the null action. 
Each action in MCTS produces an output \( s_i \in \mathcal{S} \) with state transitions defined as $s_i = L\bigl(x, s_1, \ldots, s_{i-1} \bigr)$, where $L$ represents an LLM.
This forms a reasoning trajectory \( \mathbf{t} = x \oplus s_1 \oplus \cdots \oplus s_k \), where $k$ is the maximum depth of the search tree.
The prediction for a trajectory is computed as $f(\mathbf{t, g})$, where $f$ aggregates the subtask outcomes in $\mathbf{t}$ along with a global evaluation $\mathbf{g}$.
A task-specific terminal reward is assigned based on the agreement between $f(\mathbf{t, g})$ and the simulated execution result.
We perform multiple rollouts, yielding a set of reasoning trajectories $\mathbf{T}=\left\{\mathbf{t}^1, \mathbf{t}^2, \ldots, \mathbf{t}^n\right\}$.
The cumulative rewards $R(\mathbf{t^i}) = \sum_{s \in \mathbf{t^i}} r(s)$  for these trajectories are used for weighted sampling to select the optimal trajectory $\mathbf{t^b}$.
The final prediction for $x$ is given by $f(\mathbf{t^b, g})$.

\subsection{Architecture Design}
\label{Sec3.2}
\begin{figure*}
    \centering
 \includegraphics[width=1\linewidth]{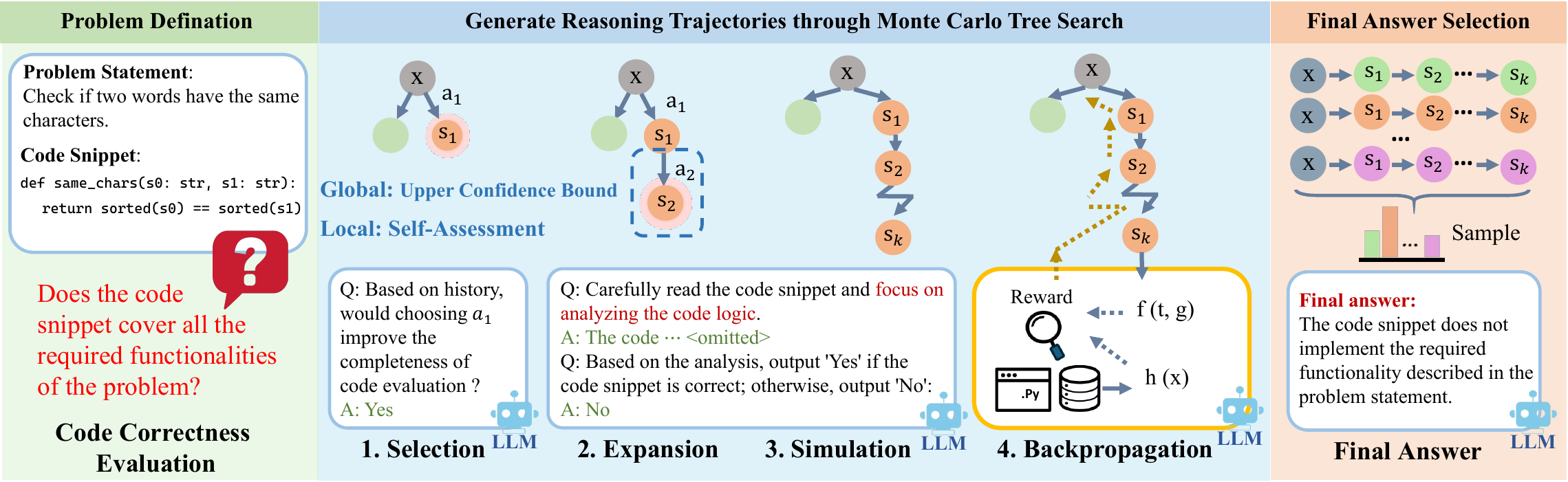}
 \vspace{-0.7cm}
 \caption{MCTS-Judge generates reasoning trajectories with multi-dimensional evaluations using Monte-Carlo Tree Search (MCTS). Each trajectory is iteratively constructed through selection, expansion, simulation, and backpropagation. Our node selection strategy combines LLM-driven self-assessment, based on historical actions in the current trajectory, with the Upper Confidence Bound for Tree (UCT) algorithm based on prior rollouts. This strategy effectively integrates global and local information, balancing the optimization of high-value regions in the search space with the refinement of the current trajectory. Moreover, we introduce a high-precision, unit-test-level reward mechanism, encouraging the LLM to perform line-by-line analysis. This simulated execution reward guides the search process and selects the final answer from candidate trajectories.
 }
\label{fig:MCTS}
\vspace{-0.3cm}
\end{figure*}

We chose MCTS to implement System-2 thinking essential for code evaluation for two reasons:
First, MCTS breaks down the overall code evaluation task into simpler subtasks, reducing the task complexity compared to other System-2 methods like Best-of-N~\cite{brown2024large} and self-consistency~\cite{wang2022self}, which require generating complete solutions in a single inference.
Second, our MCTS introduces rewards to guide the search and select the optimal trajectory, further improving the reliability of the LLM-as-a-Judge paradigm.
As shown in Fig.~\ref{fig:MCTS}, our tailored MCTS follows four key stages: selection, expansion, simulation, and backpropagation.

\textbf{1) Selection.}
The selection process begins at the root node and progresses hierarchically until it reaches a node that has not been fully expanded yet.
We propose a selection strategy that combines global and local information to balance the optimization of high-value regions in the search space with the current trajectory, resulting in a more coherent evaluation.
Specifically, we employ a two-level approach: a global-level UCT algorithm~\cite{kocsis2006bandit}, leveraging insights from previous rollouts, and a local-level LLM-driven self-assessment, which evaluates historical actions within the current trajectory.
The final selection is obtained through weighted sampling, with the UCT result weighted by $w_u$ and the self-assessment result weighted by $w_l$.
The UCT algorithm selects the node with the highest UCT value, computed as:

\vspace{-0.25cm}
\begingroup
\small
\begin{equation}
UCT(s)=  \frac{Q(s)}{N(s)}+\alpha \cdot \sqrt{\frac{\ln N_{parent}(s)}{N(s)}},
\end{equation}
\endgroup
\vspace{-0.45cm}

\noindent where $Q(s)$ represents the cumulative reward of node $s$, $N(s)$ is the visit count of $s$, $N_{parent}(s)$ the visit count of $s$'s parent node, and $\alpha$ is a constant that helps balance exploration and exploitation.
The LLM self-assessment result is obtained by prompting the LLM whether including this subtask enhances code evaluation completeness based on the completed subtasks in the current trajectory.

\textbf{2) Expansion.}
If the maximum depth has not been reached, a new child node is added to the selected node by randomly sampling an unused action and executing it.
If the action is not null, a subtask outcome is obtained by prompting the LLM to carefully analyze $c$ and $p$ (optionally with reference code) from a specific perspective and then summarize the analysis into a binary decision.

\textbf{3) Simulation.}
During the simulation process, MCTS-Judge consistently selects non-null actions to execute until the maximum depth is reached.
At this point, a complete reasoning trajectory \( \mathbf{t} = x \oplus s_1 \oplus \cdots \oplus s_k \) that evaluates the code across multiple dimensions is generated.
The prediction of this trajectory (i.e. $f(\mathbf{t, g})$) is determined by a consistency check using the majority vote across all binary subtask outcomes, combined with an additional global evaluation $g$.

\textbf{4) Backpropagation.}
Once the maximum depth is reached, a terminal reward is calculated for the trajectory and propagated upward through the search tree.
Each node in the trajectory updates its $Q(s)$ by adding the terminal reward and incrementing its $N(s)$ by one.

\subsection{Reward Mechanism}
\label{Sec3.3}
Reward is crucial in MCTS to guide the search toward promising paths while minimizing suboptimal exploration.
Moreover, cumulative rewards directly determine the final answer in MCTS-Judge, further underscoring the importance of reward accuracy.
However, verifying the correctness of predictions without ground truth labels is challenging. 
Approaches like $\mathrm{M}^*$~\cite{kang2024mindstar} and LLaMA-Berry~\cite{zhang2024llama} attempted to address this issue by training a reward model, but these methods often struggle with data collection and risk overfitting.
RAP~\cite{hao2023reasoning} introduced a self-evaluation mechanism where rewards are derived by asking the LLM to identify errors in its reasoning within a single completion. 
However, this mechanism may perform close to random if the LLM's capabilities are limited~\cite{qi2024mutual}.

\begin{figure}
    \vspace{0.1cm}
    \centering
    \includegraphics[width=0.8\linewidth]{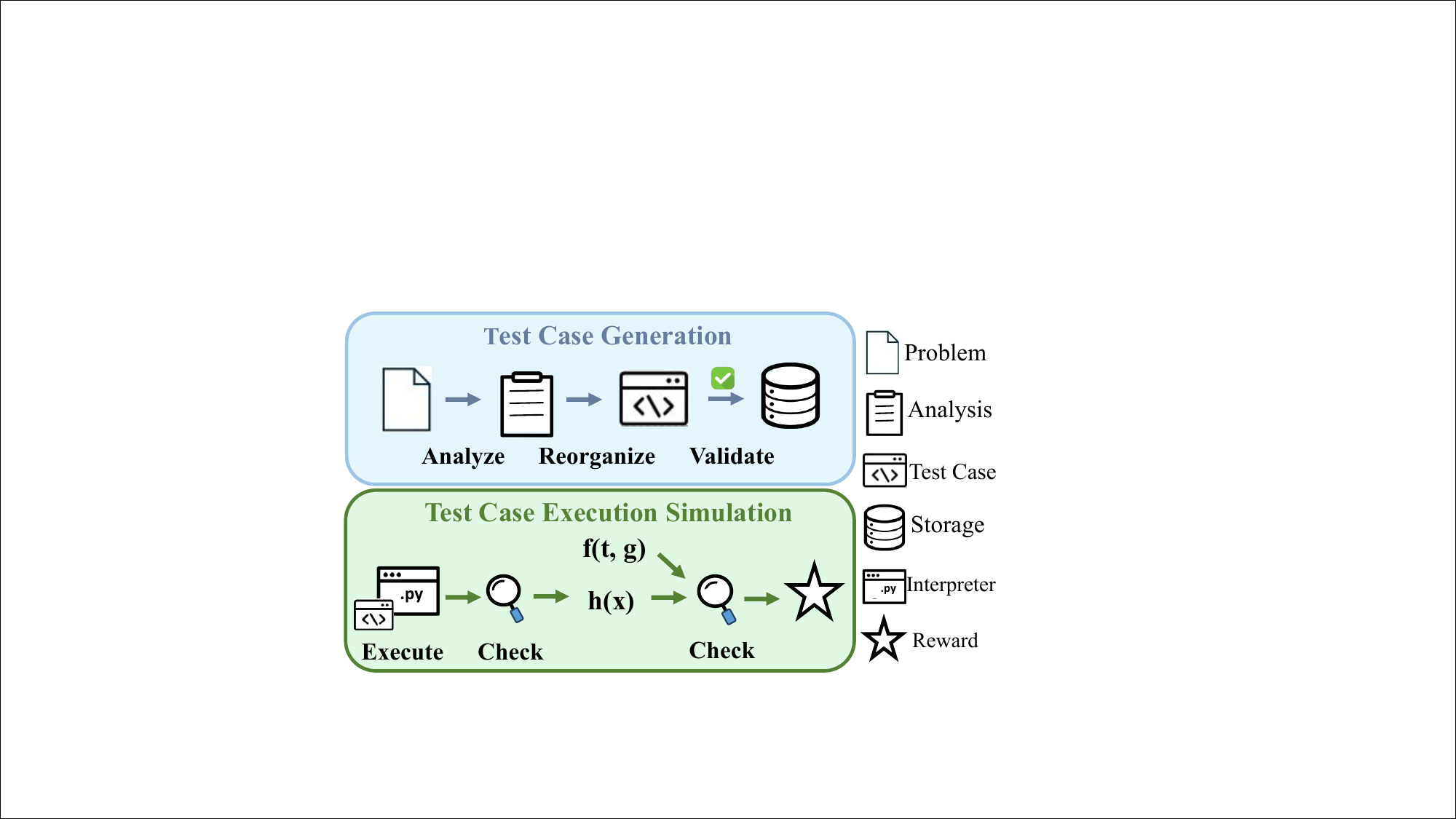}
    \vspace{-0.3cm}
    \caption{Flowchart of the fully LLM-driven Simulated Execution Reward Mechanism.  $f(\mathbf{t, g})$ represents the prediction of the trajectory, and $h(\mathbf{x})$ represents the simulated execution result.
}
    \label{fig:reward}
    \vspace{-0.3cm}
\end{figure}

Therefore, inspired by execution-based evaluation methods~\cite{liu2024marscode, xia2024agentless, zhang2024autocoderover} and the recently established success of commercial models, represented by GPT-4o, in code-related tasks and applications~\cite{ma2025dynamic, liu2025iterative, cursorai}, we propose a fully LLM-powered simulated execution reward mechanism that requires no training and improves reliability through cross-checking and step-by-step, in-depth analysis.
As illustrated in Fig.\ref{fig:reward}, the mechanism comprises two key phases: test case generation and execution simulation. The test case generation phase occurs prior to the MCTS process and requires only the problem statement $p$. GPT-4o\cite{openai2024gpt4o} is used to construct, validate, and store diverse test cases. The execution simulation phase is invoked once MCTS reaches its maximum depth and is carried out using the same LLM as in the search phase. Each test case is represented as an input-output pair, where the evaluated code must produce the correct output corresponding to a given input.

\textbf{1) Recipe for Test Case Generation.} In the test case generation phase, we instruct GPT-4o to analyze the problem statement thoroughly, identifying key requirements, constraints, boundary conditions, and special cases. Based on this analysis, the model generates low-complexity test cases that span a range of scenarios, each with a brief explanation. These cases are organized into structured input-output pairs. To ensure correctness, each pair undergoes a validation process in which the LLM is prompted $\beta$ times to self-evaluate whether the output aligns with the input and the intended behavior. Test cases that consistently pass this validation are retained, while those that do not are discarded.

\textbf{2) LLM-driven Execution Simulation.} When the MCTS search reaches its maximum depth, the test case execution simulation phase begins.
We randomly select $\gamma$ stored test cases, mask their outputs, and provide the inputs to the LLM one by one.
We instruct the LLM to simulate a code interpreter, executing the code line by line while tracking variable changes, and then determining the expected output for the given input based on this execution trace.
This process repeats $\delta$ times per test case, and the generated outputs are compared with the originally stored outputs. 
The majority vote from the $\delta$ repetitions finally determines whether a test case passes.
The reward mechanism predicts that the code is correct only if all sampled test cases pass, and the result is expressed by $h(\mathbf{x})$.
This design mirrors practical test case evaluation: if a code passes all test cases, it may be correct; however, if it fails any test case, it is definitively incorrect.
Finally, if the trajectory's prediction $f(\mathbf{t, g})$ matches $h(\mathbf{x})$, the trajectory receives a terminal reward $\epsilon$.

In doing so, our reward mechanism is both cautious and reliable, leveraging the characteristics of the code evaluation task to establish a systematic, cross-checking evaluation process that effectively minimizes errors.
Additionally, by simulating an interpreter that executes the code line by line, our approach encourages LLMs to perform fine-grained deductive reasoning, considering code flow, variable updates, and logical branches.
This detailed analysis helps uncover potential errors that might otherwise go unnoticed with a superficial ``general impression'', ensuring that final conclusions are grounded in concrete and verifiable evidence.

\section{Experiments}

\begin{table*}[h!]
    \centering
    \caption{Accuracy (\%) of MCTS-Judge and baselines on BigCodeBench, APPS, and HumanEval-X.
MCTS-Judge significantly improves the accuracy of base models and achieves the highest accuracy among existing LLM-as-a-Judge methods across all benchmarks and five LLMs (highlighted in bold).
It also surpasses larger reasoning model, \textit{Qwen-QwQ-32B} in most tasks and outperforms o1-series models in certain tasks (highlighted with underlines).}
    \label{table:main}
    \scalebox{0.83}{
    \renewcommand{\arraystretch}{0.95}
    \begin{tabular}{l c c c c c c c c c}
        \toprule
        \multirow{2}{*}{Method} 
        & \multirow{2}{*}{Approach} 
        & \multirow{2}{*}{BigCodeBench}
        & \multirow{2}{*}{APPS}
        & \multicolumn{6}{c}{HumanEval-X} \\
        \cmidrule(lr){5-10}
        & & & & Python & Java & C++ & JavaScript & Go & Average \\
        \midrule
        \multicolumn{10}{l}{\textit{Test-case Verification}} \\
        GPT-4o & System-1 & - & 55.00 & 59.09 & 53.79 & 60.61 & 62.12 & 65.91 & 60.30 \\
        \midrule
        \multicolumn{10}{l}{\textit{Commercial Reasoning LLMs}} \\
        GPT-o1-preview & System-2 & 82.02 & 75.00 & 82.58 & 89.39 & 87.12 & 86.36 & 83.33 & 85.76 \\
        GPT-o1-mini   & System-2 & 75.70 & 78.00 & 95.45 & 92.42 & 94.70 & 90.91 & 88.64 & 92.42 \\
        \midrule
        \multicolumn{10}{l}{\textit{Open-sourced Reasoning LLMs}} \\
        Qwen-QwQ-32B & System-2 & 50.96 & 60.00 & 72.73 & 75.00 & 75.00 & 64.39 & 78.03 & 73.03 \\
        \midrule
        \midrule
        \multicolumn{10}{c}{\textit{Code-Specialized Base Model: Qwen2.5-Coder-14B-Instruct}} \\
        Vanilla   & System-1 & 63.33 & 62.00 & 62.12 & 64.39 & 68.94 & 64.39 & 73.48 & 66.66 \\
        ICE-Score    & System-1 & 70.44 & 65.00 & 72.73 & 74.24 & 71.97 & 72.73 & 78.79 & 74.09 \\
        CodeJudge    & System-1 & 63.33 & 68.00 & 86.36 & 81.06 & 79.55 & 82.58 & 75.75 & 81.06 \\
        \rowcolor{gray!20} MCTS-Judge (Ours) 
            & System-2
            & \textbf{71.23} 
            & \textbf{\underline{79.00}} 
            & \textbf{\underline{90.15}}
            & \textbf{85.61} 
            & \textbf{84.09} 
            & \textbf{84.85} 
            & \textbf{81.06} 
            & \textbf{85.15} \\
        \midrule
        \multicolumn{10}{c}{\textit{Code-Specialized Base Model: DeepSeek-Coder-V2-16B-Instruct}} \\
        Vanilla  & System-1 & 51.75 & 41.00 & 73.48 & 64.39 & 63.64 & 66.67 & 60.61 & 65.76 \\
        ICE-Score    & System-1 & 57.89 & 48.00 & 71.21 & 76.52 & 69.70 & 70.45 & 74.24 & 72.42 \\
        CodeJudge    & System-1 & 52.45 & 62.00 & 73.48 & 69.70 & 67.42 & 69.70 & 66.67 & 69.39 \\
        \rowcolor{gray!20} MCTS-Judge (Ours) 
            & System-2 
            & \textbf{62.46} 
            & \textbf{\underline{80.00}} 
            & \textbf{80.30} 
            & \textbf{77.27} 
            & \textbf{80.30} 
            & \textbf{78.79} 
            & \textbf{82.58} 
            & \textbf{79.85} \\
        \midrule
        \multicolumn{10}{c}{\textit{Code-Specialized Base Model: Mistralai-Codestral-22B}} \\
        Vanilla   & System-1 & 42.81 & 62.00 & 82.58 & 68.18 & 70.45 & 62.88 & 66.67 & 70.15 \\
        ICE-Score    & System-1 & 51.93 & 56.00 & 82.58 & 68.18 & 60.61 & 63.64 & 61.36 & 67.27 \\
        CodeJudge    & System-1 & 49.04 & 54.00 & 85.61 & 69.70 & 68.94 & 71.21 & 66.67 & 72.43 \\
        \rowcolor{gray!20} MCTS-Judge (Ours) 
            & System-2
            & \textbf{68.77} 
            & \textbf{72.00} 
            & \textbf{\underline{87.78}}
            & \textbf{75.76} 
            & \textbf{77.27} 
            & \textbf{73.48} 
            & \textbf{75.76} 
            & \textbf{78.01} \\
        \midrule
        \multicolumn{10}{c}{\textit{General Base Model: Llama-3.1-8B-Instruct}} \\
        Vanilla   & System-1 & 43.16 & 56.00 & 65.91 & 63.64 & 64.39 & 62.12 & 70.45 & 65.30 \\
        ICE-Score    & System-1 & 45.88 & 42.00 & 72.73 & 64.39 & 62.88 & 56.82 & 54.55 & 62.27 \\
        CodeJudge    & System-1 & 63.86 & 53.00 & 73.48 & 73.48 & 75.76 & 70.45 & 67.42 & 72.12 \\
        \rowcolor{gray!20} MCTS-Judge (Ours) 
            & System-2 
            & \textbf{71.84} 
            & \textbf{62.00} 
            & \textbf{74.24} 
            & \textbf{79.55} 
            & \textbf{77.27} 
            & \textbf{70.45} 
            & \textbf{71.97} 
            & \textbf{74.70} \\
        \midrule
        \multicolumn{10}{c}{\textit{Commercial General Base Model: GPT-4o-mini}} \\
        Vanilla   & System-1 & 72.37 & 65.00 & 86.36 & 82.58 & 85.61 & 86.36 & 84.85 & 85.15 \\
        ICE-Score    & System-1 & 77.37 & 72.00 & 84.85 & 78.79 & 86.36 & 83.33 & 85.61 & 83.79 \\
        CodeJudge    & System-1 & 70.70 & 72.00 & 87.12 & 83.33 & 87.88 & 86.36 & 84.09 & 85.76 \\
        \rowcolor{gray!20} MCTS-Judge (Ours) 
            & System-2 
            & \textbf{\underline{79.12}}
            & \textbf{\underline{76.00}} 
            & \textbf{\underline{87.88}}
            & \textbf{86.36} 
            & \textbf{\underline{88.64}}
            & \textbf{\underline{88.64}} 
            & \textbf{\underline{85.61}}
            & \textbf{\underline{87.43}} \\
        \midrule
        \bottomrule
    \end{tabular}
    }
    \vspace{-0.5cm}
\end{table*}

\subsection{Setup}
Following previous work~\cite{tong2024codejudge}, we evaluated MCTS-Judge on three challenging benchmarks: HumanEval-X~\cite{zheng2023codegeex}, APPS~\cite{hendrycks2021measuring}, and BigCodeBench~\cite{zhuo2024bigcodebench}.
HumanEval-X includes 164 introductory coding tasks across five programming languages.
APPS consists of Python coding tasks of three different difficulty levels, from which we randomly selected 100 competition-level tasks.
BigCodeBench contains 1,140 practical and challenging Python programming tasks, covering 723 function calls from 139 libraries.
For tasks in BigCodeBench that lack meaningful input-output pairs, such as drawing or compressing, we shift the reward mechanism from simulated execution to simulated discussions, granting rewards only when all reasoning steps yield positive signals that enhance generalizability.

MCTS-Judge is a general framework compatible with various LLMs. 
To assess its effectiveness and generalizability, we employed five different LLMs as base models, including code-specialized LLMs: \textit{Qwen2.5-Coder-14B}~\cite{hui2024qwen2}, \textit{DeepSeek-Coder-V2-16B-Instruct}~\cite{zhu2024deepseek} and \textit{Mistralai-Codestral-22B}~\cite{Codestral}, as well as general LLMs:  \textit{Llama-3.1-8B-Instruct}~\cite{touvron2023llama} and \textit{GPT-4o-mini}~\cite{achiam2023gpt}.
We compare MCTS-Judge with three System 2 thinking LLMs, including OpenAI \textit{o1-preview}~\cite{jaech2024openai}, \textit{o1-mini}~\cite{jaech2024openai}, and \textit{Qwen-QwQ-32B}~\cite{qwen2024technical}, as well as two LLM-as-a-Judge paradigms designed for code evaluation with System 1 thinking: CodeJudge~\cite{tong2024codejudge} and ICE-Score~\cite{zhuo2023ice}.\footnote{ICE-Score produces ratings ranging from 0 to 4. Following the approach in~\cite{tong2024codejudge}, only a rating of 4 is considered correct.}
We further introduce a Vanilla baseline, which prompts the LLM directly for code correctness, reflecting its native evaluation capability. Additionally, we also include a test-case-only verification baseline using GPT-4o, demonstrating that the MCTS-Judge's effectiveness derives from its architecture rather than model-specific advantages. BigCodeBench is excluded here as it's evaluated with non-test-case-based approach.
More details such as hyperparameters and prompts are included in Appendix~\ref{appendix:exp-setting}.

\begin{figure}[h]
    \centering
    \includegraphics[width=0.8\linewidth]{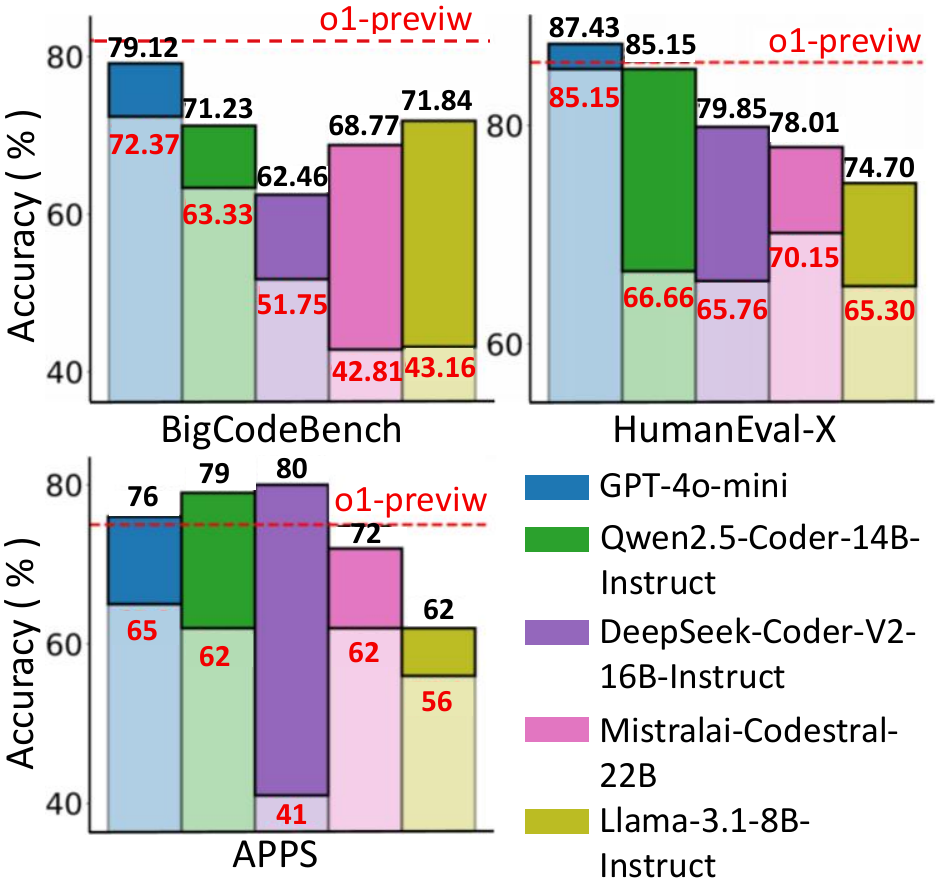}
\caption{MCTS-Judge (darker colors) significantly enhances LLMs' inherent code evaluation capabilities (lighter colors) across three benchmarks.}
\label{fig:compare}
\vspace{-0.5cm}
\end{figure}

\subsection{Main Results}
Table~\ref{table:main} presents the comparison results between MCTS-Judge and baselines.
We highlight three key observations: (1) MCTS-Judge significantly enhances the code evaluation capabilities of all base models. 
When using open-source LLMs with substantially smaller model sizes, its performance can match or even surpass o1-series models.
This phenomenon is illustrated more clearly in Fig.~\ref{fig:compare}. 
On average, MCTS-Judge achieves a 14.34\% accuracy improvement across five different base models on three benchmarks.
In particular, \textit{DeepSeek-Coder-V2-16B-Instruct}, originally at 41\% accuracy on the APPS benchmark, improved dramatically to 80\% with MCTS-Judge, surpassing both \textit{o1-preview} and \textit{o1-mini}.
(2) Any base model we evaluated, when powered by MCTS-Judge, outperforms the open-source reasoning model \textit{Qwen-QwQ-32B} on most tasks.
For instance, MCTS-Judge based on \textit{Llama-3.1-8B-Instruct}, with a model size only a quarter of \textit{Qwen-QwQ-32B}, outperforms it in all tasks except those using the Go language, achieving up to a 20.88\% higher accuracy.
(3) Compared to previous LLM-as-a-Judge paradigms with System 1 thinking, MCTS-Judge demonstrates significantly superior performance in all tasks.
For example, MCTS-Judge with \textit{DeepSeek-Coder-V2-16B-Instruct} achieved 18\% higher accuracy than CodeJudge and 32\% higher than ICE-Score on the APPS benchmark.

\subsection{Inference Efficiency}
To evaluate the test-time computational efficiency of MCTS-Judge, we analyzed the average number of reasoning tokens generated on the APPS benchmark.
As presented in Table~\ref{tab:reasoning_tokens_accuracy}, when using \textit{DeepSeek-Coder-V2-16B-Instruct} as the base model, MCTS-Judge outperforms \textit{o1-preview} in accuracy, while only consuming one third as many reasoning tokens, with an additional 20\% equivalent token consumption for simulation in parallelization indicated by the superscript, and maintaining a model size that is 19 times smaller\footnote{The model sizes of o1-preview, o1-mini, and GPT-4o-mini are referenced from this paper~\cite{abacha2024medec}.}.

\begin{table}[h!]
\centering
\renewcommand{\arraystretch}{1} 
\scalebox{0.82}{
\begin{tabular}{cp{1.7cm}p{2cm}p{0.8cm}}
\toprule
\multirow{2}{*}{Methods} & \multirow{2}{*}{Model Size} & \# Reasoning & \multirow{2}{*}{Acc} \\ 
 &  & Tokens &  \\ 
\midrule
o1-preview & \textasciitilde300B & 5631 & 75.0 \\ 
o1-mini    & \textasciitilde100B & 3755 & 78.0 \\ 
Qwen-QwQ-32B   & 32B  & 2559 & 60.0 \\ 
\rowcolor{gray!20} Ours w/ Deepseek & \textbf{16B} & \textbf{2065}\textsubscript{\footnotesize  +412} & \textbf{80.0} \\ 
\bottomrule
\end{tabular}
}
\caption{Compared to advanced reasoning LLMs, MCTS-Judge is cost-effective. With \textit{DeepSeek-Coder-V2-16B-Instruct} on the APPS benchmark, it achieves the highest accuracy using the fewest tokens and the smallest model size.}
\label{tab:reasoning_tokens_accuracy}
\vspace{-0.5cm}
\end{table}

\subsection{Fine-grained Quality Assessment}
\label{Sec:fine-grained}
MCTS-Judge demonstrated superior code correctness evaluation ability, while simultaneously generating multi-perspective analyses during reasoning trajectory construction.
We believe that this may offer developers deeper insights into the code, providing a distinct advantage over both similarity-based and execution-based evaluation methods.
To evaluate the quality of meta-analysis and reasoning capabilities, we compared the reasoning trajectories generated by MCTS-Judge with three reasoning models—\textit{o1-preview}, \textit{o1-mini}, and \textit{Qwen-QwQ-32B} across four critical dimensions: thoroughness, logic, analysis, and overall reasoning quality, with GPT-4o assessing the win rate.
As shown in Table~\ref{tab:rag_comparison}, MCTS-Judge with \textit{Deepseek-Coder-V2-16B-Instruct} and \textit{Qwen2.5-Coder-14B-Instruct} consistently achieves higher win rates, particularly excelling at thoroughness and depth of analysis.

\begin{table}[h!]
\centering
\setlength{\tabcolsep}{3pt} 
\renewcommand{\arraystretch}{0.87}
\scalebox{0.83}{
\begin{tabular}{l c c c c}
\toprule
\multirow{2}{*}{Dimensions} & \multicolumn{2}{c}{\shortstack{Deepseek-Coder-\\V2-16B-Instruct}} & \multicolumn{2}{c}{\shortstack{Qwen2.5-Coder-\\14B-Instruct}} \\
\cmidrule(lr){2-3} \cmidrule(lr){4-5}
 & o1-preview & \cellcolor{gray!20} Ours & o1-preview & \cellcolor{gray!20} Ours\\
\midrule
Thoroughness & 35.6\% & \cellcolor{gray!20} \textbf{64.4\%} & 28.8\% & \cellcolor{gray!20} \textbf{71.2\%} \\
Logic             & \textbf{51.5\%} & \cellcolor{gray!20} 48.5\% & 49.2\% & \cellcolor{gray!20} \textbf{50.8\%} \\
Analysis         & 33.3\% & \cellcolor{gray!20} \textbf{66.7\%} & 34.8\% & \cellcolor{gray!20} \textbf{65.2\%} \\
Overall           & \textbf{54.5\%} & \cellcolor{gray!20} 45.5\% & \textbf{52.3\%} & \cellcolor{gray!20} 47.7\% \\
\cmidrule(lr){2-3} \cmidrule(lr){4-5}
 & o1-mini & \cellcolor{gray!20} Ours & o1-mini & \cellcolor{gray!20} Ours\\
\midrule
Thoroughness & 25.8\% & \cellcolor{gray!20} \textbf{74.2\%} & 14.4\% & \cellcolor{gray!20} \textbf{85.6\%} \\
Logic             & 40.9\% & \cellcolor{gray!20} \textbf{59.1\%} & 22.7\% & \cellcolor{gray!20} \textbf{77.3\%} \\
Analysis          & 16.7\% & \cellcolor{gray!20} \textbf{83.3\%} & 15.9\% & \cellcolor{gray!20} \textbf{84.1\%} \\
Overall           & 46.2\% & \cellcolor{gray!20} \textbf{53.8\%} & 47.0\% & \cellcolor{gray!20} \textbf{53.0\%} \\
\cmidrule(lr){2-3} \cmidrule(lr){4-5}
 & QwQ & \cellcolor{gray!20} Ours & QwQ & \cellcolor{gray!20} Ours\\
\midrule
Thoroughness & 28.0\% & \cellcolor{gray!20} \textbf{72.0\%} & 31.1\% & \cellcolor{gray!20} \textbf{68.9\%} \\
Logic             & 25.8\% & \cellcolor{gray!20} \textbf{74.2\%} & 43.2\% & \cellcolor{gray!20} \textbf{56.8\%} \\
Analysis         & 38.6\% & \cellcolor{gray!20} \textbf{61.4\%} & 40.1\% & \cellcolor{gray!20} \textbf{59.9\%} \\
Overall           & 49.2\% & \cellcolor{gray!20} \textbf{50.8\%} & 43.9\% & \cellcolor{gray!20} \textbf{56.1\%} \\
\bottomrule
\end{tabular}
}
\caption{Comparison of MCTS-Judge's reasoning trajectories with advanced reasoning LLMs across thoroughness, logic, analysis, and overall reasoning quality, with GPT-4o assessing the win rate.}
\label{tab:rag_comparison}
\vspace{-0.5cm}
\end{table}

\subsection{Extensions to General Scenarios}
\label{sec:generlizable}
Reference code is crucial for similarity-based evaluation but is often unavailable in practice.
While it aids LLMs in understanding problems, LLM-as-a-Judge methods should adapt to more generalizable settings without it. 
We evaluated MCTS-Judge and baselines on three benchmarks without reference code (full results in Appendix~\ref{appendix:wo_ref}).
As shown in Table~\ref{tab:method_comparison}, the absence of reference code significantly degrades the performance of existing LLM-as-a-Judge frameworks.
In contrast, our MCTS-Judge demonstrates exceptional robustness, with only minimal performance drop, highlighting its promising generalization capabilities.
\begin{table}[h!]
\centering
\renewcommand{\arraystretch}{0.95}
\scalebox{0.825}{
\begin{tabular}{p{2.7cm} p{1.6cm} p{1.6cm} p{1.6cm}}
\toprule
\multirow{2}{*}{Method} & BigCode- & \multirow{2}{*}{APPS} & Human \\ 
 & Bench &  & Eval-X \\ 
\midrule
ICE-Score       
  & 45.9 
  & 42.0 
  & 62.3 \\
w/o reference     
  & 34.5 {\scriptsize(\textcolor{red}{-11.4\%})}
  & 46.0 {\scriptsize(\textcolor{ForestGreen}{+4\%})}
  & 52.3 {\scriptsize(\textcolor{red}{-10.0\%})} \\
\midrule
CodeJudge      
  & 63.9 
  & 53.0 
  & 72.1 \\
w/o reference     
  & 41.4 {\scriptsize(\textcolor{red}{-22.5\%})}
  & 47.0 {\scriptsize(\textcolor{red}{-6.0\%})}
  & 57.4 {\scriptsize(\textcolor{red}{-14.7\%})} \\
\midrule
\rowcolor{gray!20}Ours 
  & \textbf{71.8} 
  & \textbf{62.0} 
  & \textbf{74.7} \\
\rowcolor{gray!20}w/o reference     
  & 65.8 {\scriptsize(\textcolor{red}{-6\%})}
  & \textbf{62.0} {\scriptsize(\textcolor{ForestGreen}{+0.0\%})}
  & 69.5 {\scriptsize(\textcolor{red}{-5.2\%})} \\
\bottomrule
\end{tabular}
}
\caption{In the absence of reference code, MCTS-Judge with \textit{Llama-3.1-8B-Instruct} demonstrates robustness with minimal performance drop, while other baselines degrade significantly.}
\label{tab:method_comparison}
\vspace{-0.5cm}
\end{table}

\subsection{Scaling Test-time Computation}
\begin{figure}
    \centering
    \includegraphics[width=0.8\linewidth]{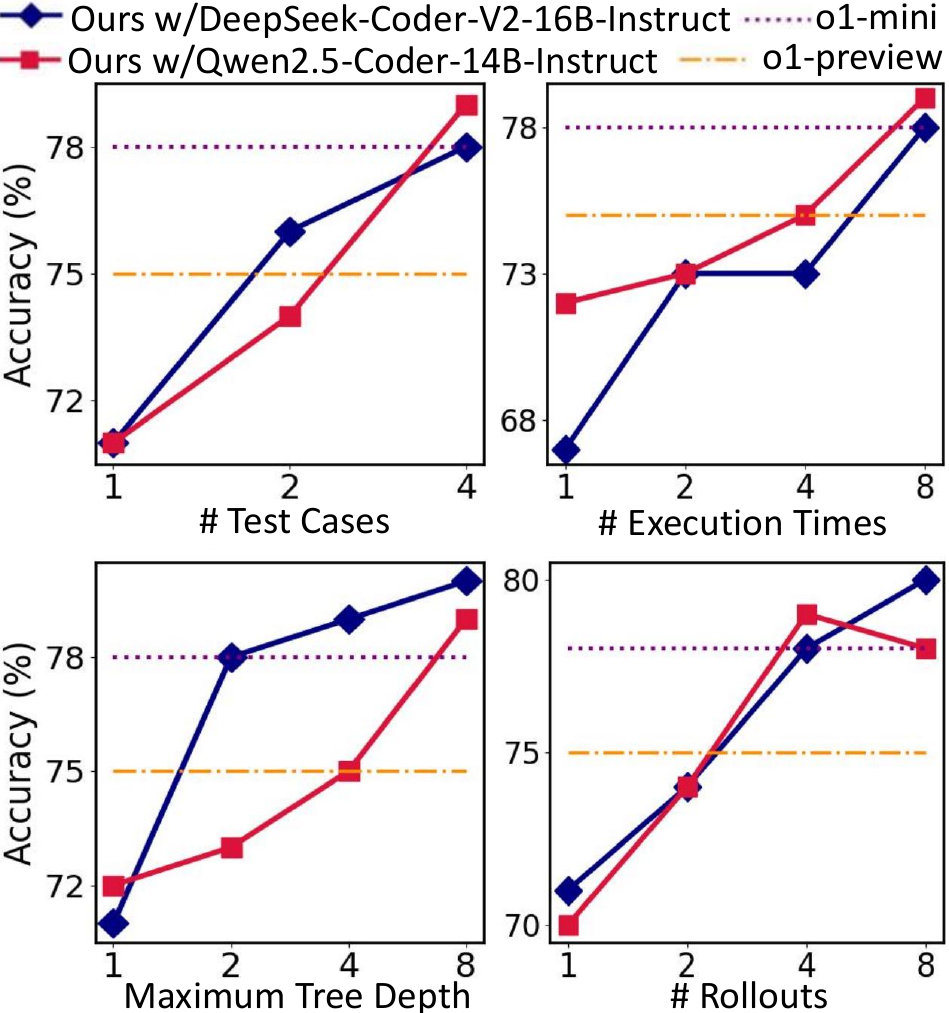}
\caption{Increasing test cases ($\alpha$), executions per case ($\delta$), tree depth, and rollouts improves MCTS-Judge's accuracy on APPS, revealing a test-time scaling law.}
    \label{fig:law}
    \vspace{-0.5cm}
\end{figure}

We explore the relationship between test-time computational scale and performance gains under our LLM-as-a-Judge framework. 
MCTS-Judge relies on simulated execution of test cases to determine the terminal reward, thereby providing more accurate guidance for MCTS and final prediction selection.
Intuitively, increasing the number of test cases ($\alpha$) reduces the likelihood of misjudging incorrect code as correct, while increasing the execution times per test case ($\delta$) further enhances accuracy.
Furthermore, extending the maximum tree depth provides a more comprehensive evaluation, and more rollouts enable broader exploration.
Fig.~\ref{fig:law} demonstrates the impact of these key hyperparameters on the APPS benchmark using \textit{DeepSeek-Coder-V2-16B-Instruct} and \textit{Qwen2.5-Coder-14B-Instruct} as base models. 
MCTS-Judge benefits from increased test-time computation, though the gains vary with specific hyperparameters and models. These results align with OpenAI’s findings~\cite{openai}, highlighting the potential of test-time scaling for LLM-as-a-Judge frameworks.

\subsection{Ablation Studies}
Table~\ref{tab:ablation_example} presents ablation results evaluating the key components of MCTS-Judge. Under System-1 thinking, the Vanilla baseline reflects the base model’s intrinsic evaluation capability, while Majority Vote executes all subtasks and selects the most frequent answer. Majority Vote improves accuracy by 13\% over Vanilla, highlighting the value of incorporating multi-perspective evaluation.

Under System-2 thinking, driven by MCTS, reward mechanisms are further analyzed. $RM_{SC}$ assigns rewards based on self-consistency majority voting~\cite{qi2024mutual}, while $RM_{SE}$ incorporates self-evaluation rewards~\cite{hao2023reasoning}. Our proposed simulated execution reward, closely aligned with ground truth, surpasses $RM_{SC}$ and $RM_{SE}$ by 13\% in accuracy.
Moreover, a variant using pure UCT-based node selection is outperformed by the full MCTS-Judge, highlighting the benefit of our global-local-aware node selection strategy.

\begin{table}[h!]
\centering
\renewcommand{\arraystretch}{0.94}
\scalebox{0.93}{
\begin{tabular}{l l l c}
\toprule
\multirow{2}{*}{Method} & Reward & Node & \multirow{2}{*}{Acc} \\
 & Model & Selection &  \\
\midrule
\multicolumn{4}{l}{\textit{System-1 Thinking}} \\[0.1cm]
Vanilla & - & - & 41.0 \\
\multirow{1}{*}{Majority Vote} & - & - & 54.0 \\
\midrule
\multicolumn{4}{l}{\textit{System-2 Thinking}} \\[0.1cm]
\multirow{4}{*}{\shortstack[l]{Monte Carlo\\Tree Search}} 
 & RM$_{SC}$   & UCT & 65.0 \\
 & RM$_{SE}$   & UCT & 65.0 \\
 & RM$_{Ours}$ & UCT & 78.0 \\
 & \cellcolor{gray!20}RM$_{Ours}$ & \cellcolor{gray!20}UCT+LLM & \cellcolor{gray!20} \textbf{80.0} \\
\bottomrule
\end{tabular}
}
\caption{Ablation of System-2 thinking, reward mechanism, and node selection strategy on APPS with \textit{DeepSeek-Coder-V2-16B-Instruct} highlights the effectiveness of our designed components. 
The grey line represents the complete MCTS-Judge, improving from 41.0\% to 80.0\% over Vanilla.}
\label{tab:ablation_example}
\vspace{-0.5cm}
\end{table}

\section{Conclusion}
In this work, we propose MCTS-Judge, a novel resource-efficient, test-time computation LLM-as-a-Judge framework with System-2 thinking for code correctness evaluation.
Powered by a fully LLM-driven MCTS, MCTS-Judge decomposes problems into simpler, multi-perspective evaluations.
Through our global-local node selection strategy, along with guidance from a simulated execution reward mechanism, MCTS-Judge performs line-by-line deep analysis.
Experiments on five LLMs and three benchmarks show that MCTS-Judge significantly improves base model accuracy, surpassing o1-series models and Qwen-QwQ-32B with one-third of the tokens and a smaller model size.
Compared to existing LLM-as-a-Judge frameworks with System-1 thinking, MCTS-Judge achieves SOTA performance while reducing dependence on reference code.
Moreover, its reasoning trajectory shows superiority in logic, analytics, thoroughness, and overall quality.
We further reveal the test-time scaling law of MCTS-Judge, marking an important first step in integrating test-time computation with the LLM-as-a-Judge paradigm.

\section{Limitations}
In this work, we propose a System-2 thinking approach with a carefully designed architecture for code correctness evaluation. Our current reward mechanism leverages GPT-4o, one of the few models capable of producing reliable and well-formatted reward signals. In contrast, existing open-source LLMs often struggle with accurate, line-by-line code execution using only pre-trained capabilities and frequently fail to generate structured, precise outputs for programming tasks. Looking forward, we aim to integrate future advancements in open-source models to develop a more cost-effective and broadly deployable solution. To ensure full and reliable reproducibility, we will release the complete codebase, data flywheel pipeline for test case generation, and comprehensive documentation upon acceptance.

\bibliography{main}

\appendix
\newpage

\section{Experiment Settings}
\label{appendix:exp-setting}
Table~\ref{table:hyperparameter} details the hyperparameter settings of MCTS-Judge employed to generate the results presented in this paper. 
These settings encompass various aspects of the MCTS architecture, including maximum tree depth, number of rollouts, exploration constant $\alpha$, LLM sample weight $w_l$, and UCT sample weight $w_u$. 
Additionally, the reward mechanism parameters include the number of test case validations $\beta$, number of test cases used $\gamma$, number of test case simulations $\delta$, and reward scaling factor $\epsilon$. 
Hyperparameters related to the LLM configuration, such as temperature, top\_p, top\_k, and maximum output tokens, are also specified. 
All experiments were executed on a single H100 GPU with 80GB of memory, ensuring consistency and reproducibility in computational performance.

\begin{table}[h]
\begin{tabular}{l|l}
\hline
Hyperparameter                         & Value \\ \hline
Maximum tree depth                     & 9     \\
Number of rollouts                     & 8     \\
Constant $\alpha$                      & 3     \\
LLM sample weight $w_l$                & 0.1   \\
UCT sample weight $w_u$                & 0.9   \\
Number of test case validations $\beta$& 5     \\
Number of test cases used $\gamma$     & 3     \\
Number of test case simulations $\delta$ & 7   \\
Reward $\epsilon$                      & 1.1   \\
Temperature                            & 0.4   \\
Top\_p                                  & 0.95  \\
Top\_k                                  & 40    \\
Maximum tokens                         & 2048  \\ \hline
\end{tabular}
\caption{Hyperparameters of MCTS-Judge.}
\label{table:hyperparameter}
\end{table}

\section{Limitations of Execution-Free Methods} 
\label{appendix:comparision-exec}
\begin{figure}
\centering
\begin{subfigure}[b]{0.23\textwidth}
    \centering
    \includegraphics[width=\textwidth]{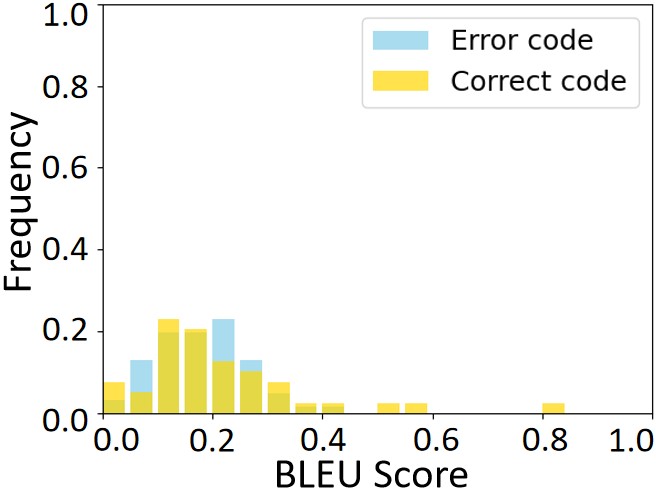}
\end{subfigure}
\begin{subfigure}[b]{0.23\textwidth}
    \centering
    \includegraphics[width=\textwidth]{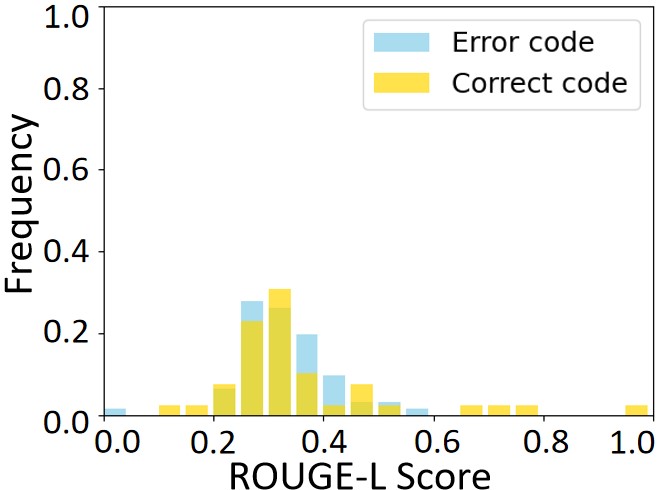}
\end{subfigure}
\begin{subfigure}[b]{0.23\textwidth}
    \centering
    \includegraphics[width=\textwidth]{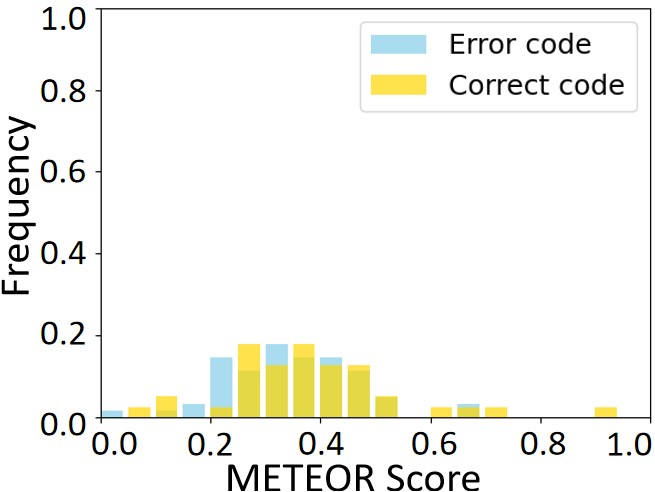}
\end{subfigure}
\centering
\begin{subfigure}[b]{0.23\textwidth}
    \centering
    \includegraphics[width=\textwidth]{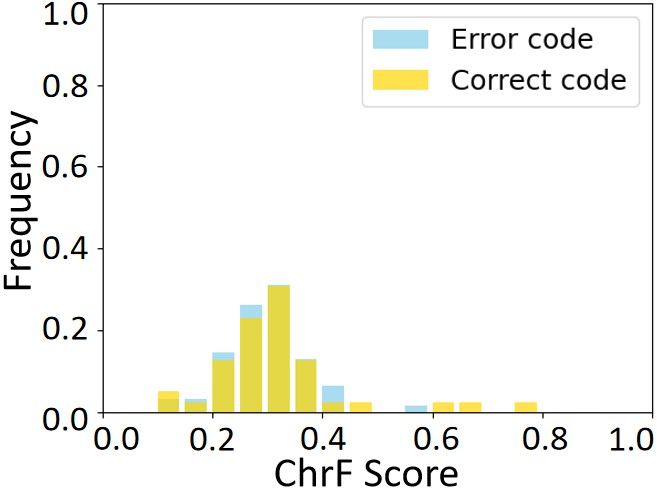}
\end{subfigure}
\begin{subfigure}[b]{0.23\textwidth}
    \centering
    \includegraphics[width=\textwidth]{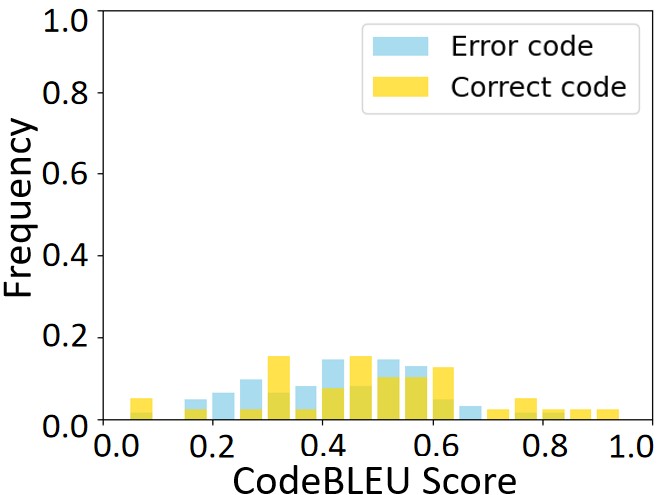}
\end{subfigure}
\begin{subfigure}[b]{0.23\textwidth}
    \centering
    \includegraphics[width=\textwidth]{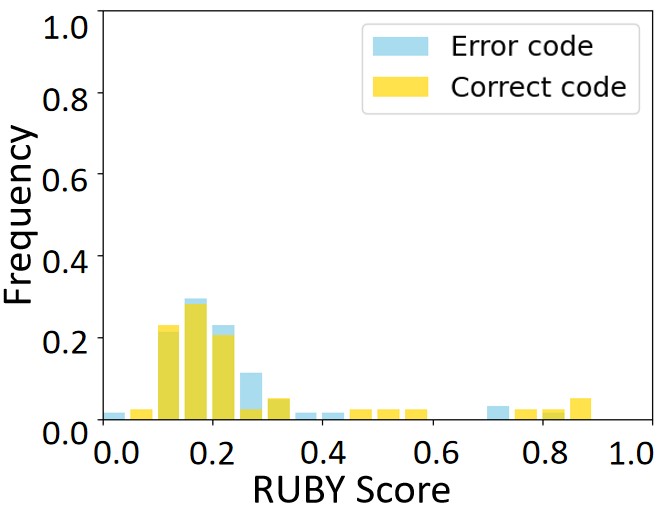}
\end{subfigure}
\caption{Score distributions of six execution-free metrics on the APPS benchmark show negligible distinction between correct and incorrect code, indicating a lack of reliability.}
\label{fig:scores}
\end{figure}

In this experiment, we evaluate whether similarity-based execution-free metrics, which do not require test cases and isolated environments, can be used to accurately assess code correctness.
We evaluated six representative metrics on the APPS benchmark: BLEU, ROUGE-L, METEOR, ChrF, CodeBLEU, and RUBY.
Fig.~\ref{fig:scores} shows that the score distributions for incorrect and correct code differ only slightly for these metrics.
This further highlights the importance of developing high-precision execution-free methods for assessing code correctness.

\section{Evaluation without References}
\label{appendix:wo_ref}
As discussed in Sec.\ref{sec:generlizable}, we present the complete results for extending to more general scenarios, specifically in the absence of reference data. Table\ref{table:wo_ref_full} compares the performance of MCTS-Judge with baselines across three benchmarks and five LLMs. The results highlight MCTS-Judge's greater robustness and generalization ability, as evidenced by reduced performance decay compared to other LLM-as-a-Judge methods, aligning with the conclusions drawn in the main paper's analysis.

\begin{table*}[h!]
    \centering
    \scalebox{0.83}{
    \renewcommand{\arraystretch}{0.95}
    \begin{tabular}{l c c c c c c c c c}
        \toprule
        \multirow{2}{*}{Method} 
        & \multirow{2}{*}{Approach} 
        & \multirow{2}{*}{BigCodeBench}
        & \multirow{2}{*}{APPS}
        & \multicolumn{6}{c}{HumanEval-X} \\
        \cmidrule(lr){5-10}
        & & & & Python & Java & C++ & JavaScript & Go & Average \\
        \midrule
        \multicolumn{10}{l}{\textit{Commercial Reasoning LLMs}} \\
        GPT-o1-mini  & System-2 & 71.14 & 78.00 & 93.18 & 86.36 & 86.36 & 89.39 & 87.88 & 88.63 \\
        \midrule
        \multicolumn{10}{l}{\textit{Open-sourced Reasoning LLMs}} \\
        Qwen-QwQ-32B & System-2 & 51.05 & 69.00 & 61.36 & 68.18 & 66.67 & 67.42 & 77.27 & 68.18 \\
        \midrule
        \midrule
        \multicolumn{10}{c}{\textit{Code-Specialized Base Model: Qwen2.5-Coder-14B-Instruct}} \\
        Base Model   & System-1 & 46.84 & 61.00 & 60.61 & 62.88 & 68.94 & 65.91 & 76.52 & 66.97 \\
        ICE-Score    & System-1 & 66.23 & 69.00 & 75.00 & 66.67 & 71.21 & 72.73 & 75.00 & 72.12 \\
        CodeJudge    & System-1 & 47.02 & 62.00 & \textbf{87.12} & 75.76 & 79.55 & 81.06 & 73.48 & 79.40 \\
        \rowcolor{gray!20} MCTS-Judge (Ours) 
            & System-2 
            & \textbf{70.35} 
            & \textbf{74.00} 
            & 86.36
            & \textbf{\underline{87.12}} 
            & \textbf{80.30} 
            & \textbf{82.58} 
            & \textbf{82.58} 
            & \textbf{83.79} \\
        \midrule
        \multicolumn{10}{c}{\textit{Code-Specialized Base Model: DeepSeek-Coder-V2-16B-Instruct}} \\
        Base Model   & System-1 & 35.09 & 49.00 & 69.70 & 58.33 & 53.79 & 53.79 & 56.06 & 58.33 \\
        ICE-Score    & System-1 & 47.19 & 56.00 & 74.24 & 69.70 & 71.21 & 66.67 & 69.70 & 70.30 \\
        CodeJudge    & System-1 & 35.44 & 62.00 & 75.76 & 66.67 & 63.64 & 66.67 & 66.67 & 67.89 \\
        \rowcolor{gray!20} MCTS-Judge (Ours) 
            & System-2
            & \textbf{51.58} 
            & \textbf{71.00} 
            & \textbf{82.58} 
            & \textbf{72.73} 
            & \textbf{84.85} 
            & \textbf{75.00} 
            & \textbf{81.82} 
            & \textbf{79.40} \\
        \midrule
        \multicolumn{10}{c}{\textit{Code-Specialized Base Model: Mistralai-Codestral-22B}} \\
        Base Model   & System-1 & 32.63 & 43.00 & 75.76 & 58.33 & 58.33 & 57.58 & 58.33 & 61.67 \\
        ICE-Score    & System-1 & 31.05 & 41.00 & 76.52 & 57.58 & 53.03 & 59.09 & 53.79 & 60.00 \\
        CodeJudge    & System-1 & 35.44 & 48.00 & 78.79 & 59.09 & 59.09 & 60.61 & 63.64 & 64.24 \\
        \rowcolor{gray!20} MCTS-Judge (Ours) 
            & System-2 
            & \textbf{52.37} 
            & \textbf{74.00} 
            & \textbf{79.55} 
            & \textbf{72.73} 
            & \textbf{75.76} 
            & \textbf{65.91} 
            & \textbf{77.27} 
            & \textbf{74.24} \\
        \midrule
        \multicolumn{10}{c}{\textit{General Base Model: Llama-3.1-8B-Instruct}} \\
        Base Model   & System-1 & 33.95 & 58.00 & 59.85 & 60.61 & 62.88 & 64.39 & 67.42 & 63.30 \\
        ICE-Score    & System-1 & 34.47 & 46.00 & 62.12 & 51.52 & 50.76 & 54.55 & 42.42 & 52.27 \\
        CodeJudge    & System-1 & 41.40 & 47.00 & 61.36 & 60.61 & 58.33 & 54.55 & 52.27 & 57.42 \\
        \rowcolor{gray!20} MCTS-Judge (Ours) 
            & System-2 
            & \textbf{65.79} 
            & \textbf{62.00} 
            & \textbf{71.21} 
            & \textbf{69.70} 
            & \textbf{65.15} 
            & \textbf{68.18} 
            & \textbf{73.48} 
            & \textbf{69.54} \\
        \midrule
        \multicolumn{10}{c}{\textit{Commercial General Base Model: GPT-4o-mini}} \\
        Base Model   & System-1 & 71.05 & 61.00 & 81.82 & 80.30 & 83.33 & 83.33 & 83.33 & 82.42 \\
        ICE-Score    & System-1 & 58.60 & 62.00 & 78.03 & 75.00 & 75.76 & 81.06 & 82.58 & 78.49 \\
        CodeJudge    & System-1 & 57.46 & 70.00 & 81.06 & 82.58 & 85.61 & 84.85 & 83.33 & 83.49 \\
        \rowcolor{gray!20} MCTS-Judge (Ours) 
            & System-2 
            & \textbf{\underline{73.68}} 
            & \textbf{72.00} 
            & \textbf{82.58} 
            & \textbf{83.33} 
            & \textbf{\underline{87.12}} 
            & \textbf{87.12} 
            & \textbf{85.61} 
            & \textbf{85.15} \\
        \midrule
        \bottomrule
    \end{tabular}
    }

\caption{Accuracy (\%) of MCTS-Judge and baselines \textbf{without reference code} on BigCodeBench, APPS, and HumanEval-X.
Compared to existing LLM-as-a-Judge methods, MCTS-Judge without reference code still achieves the highest accuracy across all benchmarks and five LLMs (highlighted in bold).
MCTS-Judge without reference code maintains its advantage over advanced reasoning LLMs, with superior accuracy compared to the o1-series models highlighted with underlines.}
    \label{table:wo_ref_full}
\end{table*}

\section{Case Study}
We demonstrated the superiority of analysis metadata from reasoning chains generated by MCTS-Judge across four key dimensions in Section~\ref{Sec:fine-grained}. To further highlight MCTS-Judge's advantages over the System-1 thinking approach in terms of accuracy and comprehensive analysis, we present a case study comparing the vanilla \textit{Deepseek-Coder-V2-16B-Instruct} model with the one integrated with MCTS-Judge.

\begin{figure}
    \centering
    \includegraphics[width=\linewidth]{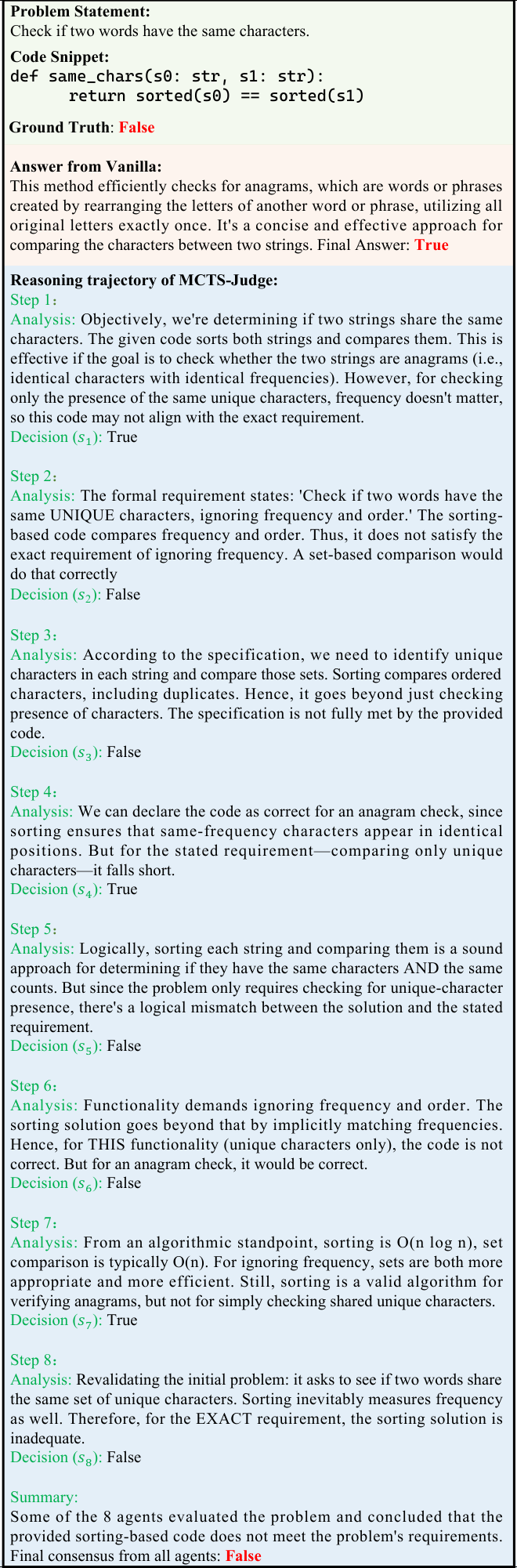}
    \caption{Case Study: System-2 Thinking vs. System-1 Thinking. MCTS-Judge leads with comprehensive evaluations, while vanilla fails with superficial analysis.}
    \label{fig:case}
\end{figure}

As shown in Fig.~\ref{fig:case}, at each step, MCTS-Judge provides evaluations from diverse perspectives, such as verifying whether requirements are met (Step 2) and assessing the correctness of the code logic (Step 5). In contrast, the vanilla model generates an incorrect answer with only superficial analysis, demonstrating a lack of deeper understanding.

\section{Prompts}
We present the key prompt designs utilized in MCTS-Judge, including the vanilla baseline (Fig.~\ref{fig:prompt1}), the test-case verification baseline (Fig.~\ref{fig:prompt6}), the LLM-driven self-assessment (Fig.~\ref{fig:prompt2}), the logic assessment action (Fig.~\ref{fig:prompt3}), the test case generation and validation (Fig.~\ref{fig:prompt4}), and the simulated execution (Fig.~\ref{fig:prompt5}).

\section{Responsible Use of Artifacts}
\label{app:responsible-artifacts}

Our experiments use publicly available research artifacts, including APPS~\citep{hendrycks2021measuring}, HumanEval-X~\citep{zheng2023codegeex}, BigCodeBench~\citep{zhuo2024bigcodebench}, public or open-weight language models, and prior code-evaluation baselines. We cite the original creators and use these artifacts only for research evaluation and comparison, following their respective licenses, access conditions, and terms of use. We do not redistribute original benchmark data, proprietary model outputs, or model weights beyond what is permitted. The study does not collect human-subject data or personally identifying information.

\begin{table*}[]
\centering
\begin{tabular}{l}
\rowcolor[HTML]{C0C0C0} 
\textbf{Vanilla}\\
\begin{tabular}[c]{@{}l@{}}\textbf{\textless{}|system|\textgreater{}}:Determine the correctness of the code snippet. Output Yes or No.\\ \textbf{\textless{}|user|\textgreater{}}: Problem Statement:\textcolor{blue}{\{problem\}}\\ Example:\textcolor{blue}{\{example\}}\\ Reference Solution (\textcolor{blue}{\{language\}}):\textcolor{blue}{\{reference code\}}\\ Code Snippet (\textcolor{blue}{\{language\}}):\textcolor{blue}{\{code\}}\\ 
\textbf{\textless{}|assistant|\textgreater{}}:Answer (Yes or No only):\end{tabular}
\end{tabular}
\caption{Prompt of the baseline named Vanilla.}
\label{fig:prompt1}
\end{table*}

\begin{table*}[]
\centering
\begin{tabular}{l}
\rowcolor[HTML]{C0C0C0} 
\textbf{Test Cases Generation}\\
\begin{tabular}[c]{@{}l@{}}\textbf{\textless{}|system|\textgreater{}}:You are an AI assistant specialized in analyzing Python functions and generating test cases. \\ You will be provided with a problem statement and its solution to evaluate the correctness of the code.\\ 
\textbf{\textless{}|user|\textgreater{}}: Full Code:\textcolor{blue}{\{language\}}\textcolor{blue}{\{code\}} \\
Public Test Case for the Main \\
Function:\textcolor{blue}{\{example\}} \\
Instructions: Please analyze how the \textcolor{blue}{\{function name\}} function is used within the main function and how \\ it contributes to the expected outputs in the gold test case. For each test case, you should analyze \\ step-by-step based on both the input and the expected output of the main function, and then provide the \\ corresponding input and expected output for the \textcolor{blue}{\{function name\}} function. Ensure that the generated test \\ cases are consistent with the behavior expected in the public test cases.\\ 
\textbf{\textless{}|assistant|\textgreater{}}:Let's break down the code and give the input and expected output of the \textcolor{blue}{\{function name\}} \\ function step-by-step for each given gold test case ignoring any discrepancie between the function's logic \\ and the expected outputs of public test case:\end{tabular} \\

\rowcolor[HTML]{C0C0C0} 
\textbf{Test Case Validation} \\
\begin{tabular}[c]{@{}l@{}}\textbf{\textless{}|system|\textgreater{}}: You are an AI assistant specializing in test case validation. Your task is to assess the \\ correctness of a given test case based on the problem description, which consists of four parts: the \\problem statement, input description, output description, and several input-output pair examples.\\ 
\textbf{\textless{}|user|\textgreater{}}: Problem:\textcolor{blue}{\{problem\}}\\ Test Cases:\textcolor{blue}{\{test\_cases\}}\\ \#\#\# Instructions: Validate the correctness of the test case and determine if it aligns with the expected \\ behavior outlined in the problem description. Do not provide a corrected version. Return PASS or FAIL\\ to indicate the accuracy of the given test case.\\ 
\textbf{\textless{}|assistant|\textgreater{}}: Let's validate the correctness of the test case step-by-step:\end{tabular}\\

\rowcolor[HTML]{C0C0C0} 
\textbf{Simulated Execution}\\
\begin{tabular}[c]{@{}l@{}}\textbf{\textless{}|system|\textgreater{}}: You are an AI assistant skilled in executing \textcolor{blue}{\{language\}} scripts.\\ 
\textbf{\textless{}|user|\textgreater{}}:Execute the following \textcolor{blue}{\{language\}} script. Analyze the code and run each subfunction in the test \\ case step-by-step.\\ Code:\textcolor{blue}{\{code\}}\\ 
Test Case: \textcolor{blue}{\{test case\}} \\ 
Instruction: Act as a \textcolor{blue}{\{language\}} interpreter to execute the code line-by-line, tracking changes in each \\ variable throughout the process. Based on this execution trace, determine the output of the unit test case.\\ 
\textbf{\textless{}|assistant|\textgreater{}}:Let's execute the code step-by-step, analyzing the input test case:\end{tabular}       \end{tabular}
\caption{Prompt for the test-case verification baseline.}
\label{fig:prompt6}
\end{table*}

\begin{table*}[]
\centering
\begin{tabular}{l}
\rowcolor[HTML]{C0C0C0} 
\textbf{LLM-drive Self-Assessment for Node Selection}\\
\begin{tabular}[c]{@{}l@{}}\textbf{\textless{}|system|\textgreater{}}: You are a code evaluation planning expert. Your task is to assess whether the suggested \\evaluator agent should proceed based on the provided problem statement, code snippet, and evaluation \\history. Determine if this evaluator will enhance coverage or completeness of the assessment.\\ 
\textbf{\textless{}|user|\textgreater{}}: Problem Statement:
\textcolor{blue}{\{problem\}}\\ Code Snippet:\textcolor{blue}{\{language\}\{code\}}\\ Proposed Next Evaluator: \textcolor{blue}{\{agent\}}\\ Evaluation History: Agents previously used: \textcolor{blue}{\{history\}}\\ Instruction: Skip the evaluation only if it very negatively affect the assessment. Otherwise, please \\respond 'Yes' to include the evaluator.\\ 
\textbf{\textless{}|assistant|\textgreater{}}: Decision: (Yes or No only)\end{tabular}
\end{tabular}
\caption{Prompt for LLM-driven self-assessment in MCTS node selection.}
\label{fig:prompt2}
\end{table*}

\begin{table*}[]
\centering
\begin{tabular}{l}
\rowcolor[HTML]{C0C0C0} 
\textbf{Code Logic Evaluation}\\
\begin{tabular}[c]{@{}l@{}}\textbf{\textless{}|system|\textgreater{}}: You will be provided with a problem statement, a code snippet that supposedly addresses \\the problem in \textcolor{blue}{\{language\}}, and a reference solution in \textcolor{blue}{\{language\}}. Your task is to check if the code \\snippet covers the required functionalities. Do not provide a corrected version.\\ Evaluation Steps:1. Read the problem statement carefully and identify the required functionalities of the \\implementation. You can refer to the example and reference answer to understand the problem better.\\2. Read the code snippet and analyze its logic. Check if the code snippet covers all the required \\functionalities of the problem. 3. Finally, conclude your evaluation.\\ 
\textbf{\textless{}|user|\textgreater{}}:Problem Statement:\textcolor{blue}{\{problem\}}\\ Example:\textcolor{blue}{\{example\}}\\ Reference Solution (\textcolor{blue}{\{language\}}):\textcolor{blue}{\{reference code\}}\\ Code Snippet (\textcolor{blue}{\{language\}}):\textcolor{blue}{\{code\}}\\ 
\textbf{\textless{}|assistant|\textgreater{}}:Evaluation (Code Logic Analysis):\end{tabular} \\
\rowcolor[HTML]{C0C0C0} 
\textbf{Analysis Summarization}\\
\begin{tabular}[c]{@{}l@{}}\textbf{\textless{}|system|\textgreater{}}: You will be provided with an analysis result of a code snippet. If the analysis believes that \\the code snippet is correct, output: "Yes". Otherwise, output: "No".\\ 
\textbf{\textless{}|user|\textgreater{}}: Analysis Result:\textcolor{blue}{\{analysis\}}\\ 
\textbf{\textless{}|assistant|\textgreater{}}: Final Answer (Yes or No only):\end{tabular}                \end{tabular}
\caption{Example of subtasks: prompt focused on logic assessment}
\label{fig:prompt3}
\end{table*}

\begin{table*}[]
\centering
\begin{tabular}{l}
\rowcolor[HTML]{C0C0C0} 
\textbf{Initial Test Cases Generation}\\
\begin{tabular}[c]{@{}l@{}}\textbf{\textless{}|system|\textgreater{}}: You are an AI assistant specializing in problem analysis and test case generation, with \\particular expertise in \textcolor{blue}{\{difficulty\}} test cases. Your task is to generate comprehensive test cases based \\on the given problem description.\\ \textbf{\textless{}|user|\textgreater{}}: Problem:\textcolor{blue}{\{problem\}}\\ \#\#\# Instructions: Please analyze the problem statement carefully and create five well-rounded test \\cases. The test cases should be at \textcolor{blue}{\{difficulty\}} difficulty level. Your test cases should: 1. Cover a \\variety of scenarios to thoroughly validate code correctness 2. Avoid excessively large computations\\3. Avoid extreme edge cases unless specifically required 4. Include common use cases and reasonable\\ boundary conditions 5. Focus on practical, real-world scenarios\\ For each test case, please provide:- Input values - Expected output- Brief explanation of what the \\test case verifies.\\ 
\textbf{\textless{}|assistant|\textgreater{}}:I'll analyze the problem systematically and create carefully curated test cases.\\ **Analysis Approach:**1. First, I'll identify the key requirements and constraints 2. Then, I'll \\determine important edge cases and boundary conditions 3. Finally, I'll design test cases that \\progressively increase in complexity\\ Let's examine each test case:\end{tabular} \\
\rowcolor[HTML]{C0C0C0} 
\textbf{JSON Reformation} \\
\begin{tabular}[c]{@{}l@{}}\textbf{\textless{}|system|\textgreater{}}: You are an AI assistant specializing in test case reformatting. Your task is to extract \\and reformat the test cases based on the examples provided.\\ \textbf{\textless{}|user|\textgreater{}}: Problem:\textcolor{blue}{\{problem\}}\\ Test Cases:\textcolor{blue}{\{test\_cases\}}\\ \#\#\# Instructions:Reformat the test cases and return them in a list of JSON format, where each test \\case is structured as follows:\\ \{\{"\textless{}root\_function\_name\textgreater{}": \{\{ "input": "\textless{}input\textgreater{}" (as a string), "expected\_output": "\textless{}expected\_\\output\textgreater{}" (as a string)\}\}\}\}\\ Use the main function name identified in each test case as the key. Retain the original input and \\output formats, ensuring that all provided test cases are included.\\ 
\textbf{\textless{}|assistant|\textgreater{}}: Reformatted Test Cases only, without explanation:\end{tabular}\\
\rowcolor[HTML]{C0C0C0} 
\textbf{Test Case Validation} \\
\begin{tabular}[c]{@{}l@{}}\textbf{\textless{}|system|\textgreater{}}: You are an AI assistant specializing in test case validation. Your task is to assess the \\correctness of a given test case based on the problem description.\\ \textbf{\textless{}|user|\textgreater{}}: Problem:\textcolor{blue}{\{problem\}}\\ Test Case:\textcolor{blue}{\{test\_case\}}\\ \#\#\# Instructions: Validate the correctness of the test case and determine if it aligns with the \\expected behavior outlined in the problem description. Do not provide a corrected version. Return \\PASS or FAIL to indicate the accuracy of the given test case.\\ \textbf{\textless{}|assistant|\textgreater{}}: Let's validate the correctness of the test case step-by-step:\end{tabular}       \end{tabular}
\caption{Prompt for test case generation and validation.}
\label{fig:prompt4}
\end{table*}

\begin{table*}[]
\begin{tabular}{l}
\rowcolor[HTML]{C0C0C0} 
\textbf{Simulated Execution}\\
\begin{tabular}[c]{@{}l@{}}\textbf{\textless{}|system|\textgreater{}}: You are an AI assistant skilled in executing \textcolor{blue}{\{language\}} scripts.\\ \textbf{\textless{}|user|\textgreater{}}: Execute the following \textcolor{blue}{\{language\}} script. Analyze the code and run each subfunction in the \\test case step-by-step.\\ Code:\textcolor{blue}{\{code\}}\\ Test Case:\textcolor{blue}{\{test\_case\}}\\ Instruction: Act as a \textcolor{blue}{\{language\}} interpreter to execute the code line-by-line, tracking changes in each \\variable throughout the process. Based on this execution trace, determine the output of the unit test case.\\ 
\textbf{\textless{}|assistant|\textgreater{}}: Let's execute the code step-by-step, analyzing the input test case:\end{tabular}                   \\
\rowcolor[HTML]{C0C0C0} 
\textbf{Compare Execution Results with Expected Outputs of Test Cases} \\        \begin{tabular}[c]{@{}l@{}}\textbf{\textless{}|system|\textgreater{}}: You are an expert in code validation with a focus on comparing code execution outputs. \\Your goal is to determine if the actual output matches the expected behavior, being flexible about \\formatting and minor differences that don't affect correctness.\\ 
\textbf{\textless{}|user|\textgreater{}}: Problem: \textcolor{blue}{\{problem\}}\\ Test Case:\textcolor{blue}{\{test\_case\}}\\ Actual Output:\textcolor{blue}{\{answer\}}\\ \#\#\# Instruction: 1. Extract the final execution result from the actual output. Compare this result with \\the expected answer, focusing on correctness rather than specific formatting. Note: Do not judge \\the code; simply validate whether the results match.\\ 
\textbf{\textless{}|assistant|\textgreater{}}: Let's check the results, and conclude with **MATCH** or **NOT MATCH** after \\the comparison.\end{tabular}
\end{tabular}
\caption{Prompt for simulated execution and verify whether the execution results match the expected answers.}
\label{fig:prompt5}
\end{table*}

\end{document}